\documentclass[sigconf]{acmart}
\usepackage{booktabs} 
\usepackage[flushleft]{threeparttable}
\usepackage{multirow}
\usepackage{bbm}
\usepackage{subcaption}
\usepackage{bm}
\usepackage{physics}
\usepackage{balance}
\usepackage{amsmath}
\usepackage[linesnumbered,boxed,ruled,vlined]{algorithm2e}
\newcommand{\eat}[1]{}

\begin{document}
\title{Predicting Path Failure In Time-Evolving Graphs}

\copyrightyear{2019} 
\acmYear{2019} 
\setcopyright{acmcopyright}
\acmConference[KDD '19]{The 25th ACM SIGKDD Conference on Knowledge Discovery and Data Mining}{August 4--8, 2019}{Anchorage, AK, USA}
\acmBooktitle{The 25th ACM SIGKDD Conference on Knowledge Discovery and Data Mining (KDD '19), August 4--8, 2019, Anchorage, AK, USA}
\acmPrice{15.00}
\acmDOI{10.1145/3292500.3330847}
\acmISBN{978-1-4503-6201-6/19/08}

\author{Jia Li}
\affiliation{%
  \institution{The Chinese University of Hong Kong}
}
\email{lijia@se.cuhk.edu.hk}

\author{Zhichao Han}
\affiliation{%
  \institution{The Chinese University of Hong Kong}
}
\email{zchan@se.cuhk.edu.hk}

\author{Hong Cheng}
\affiliation{%
  \institution{The Chinese University of Hong Kong}
}
\email{hcheng@se.cuhk.edu.hk}

\author{Jiao Su}
\affiliation{%
  \institution{The Chinese University of Hong Kong}
}
\email{jiaosu@se.cuhk.edu.hk}

\author{Pengyun Wang}
\affiliation{%
  \institution{Noah's Ark Lab, Huawei Technologies}
}
\email{wangpengyun@huawei.com}

\author{Jianfeng Zhang}
\affiliation{%
  \institution{Noah's Ark Lab, Huawei Technologies}
}
\email{zhangjianfeng3@huawei.com}

\author{Lujia Pan}
\affiliation{%
  \institution{Noah's Ark Lab, Huawei Technologies}
}
\email{panlujia@huawei.com}

\renewcommand{\shortauthors}{J. Li et al.}

\begin{abstract}
In this paper we use a time-evolving graph which consists of a sequence of graph snapshots over time to model many real-world networks.  We study the path classification problem in a time-evolving graph, which has many applications in real-world scenarios, for example, predicting path failure in a telecommunication network and predicting path congestion in a traffic network in the near future.  In order to capture the temporal dependency and graph structure dynamics, we design a novel deep neural network named Long Short-Term Memory R-GCN (LRGCN).  LRGCN considers temporal dependency between time-adjacent graph snapshots as a special relation with memory, and uses relational GCN to jointly process both intra-time and inter-time relations.  We also propose a new path representation method named \underline{s}elf-\underline{a}ttentive \underline{p}ath \underline{e}mbedding (SAPE), to embed paths of arbitrary length into fixed-length vectors.  Through experiments on a real-world telecommunication network and a traffic network in California, we demonstrate the superiority of LRGCN to other competing methods in path failure prediction, and prove the effectiveness of SAPE on path representation.

\end{abstract}

%
%

\ccsdesc[500]{Mathematics of computing~Graph algorithms}
\ccsdesc[500]{Computing methodologies~Supervised learning by classification}

\keywords{time-evolving graph; path representation; classification}


\maketitle

\section{Introduction}


Graph has been widely used to model real-world entities and the relationship among them.  For example, a telecommunication network can be modeled as a graph where a node corresponds to a switch and an edge represents an optical fiber link; a traffic network can be modeled as a graph where a node corresponds to a sensor station and an edge represents a road segment.  In many real scenarios, the graph topological structure may evolve over time, e.g., link failures due to hardware outages; road closures due to accidents or natural disasters.  This leads to a \emph{time-evolving graph} which consists of a sequence of graph snapshots over time.  In the literature some studies on time-evolving graphs focus on the node classification task, e.g., \cite{aggarwal2011node} uses a random walk approach to combine structure and content for node classification, and \cite{gunecs2014ga} improves the performance of node classification in time-evolving graphs by exploiting genetic algorithms.  In this work, we focus on a more challenging but practically useful task: \emph{path classification in a time-evolving graph}, which predicts the status of a path in the near future.  A good solution to this problem can benefit many real-world applications, e.g., predicting path failure (or path congestion) in a telecommunication (or traffic) network so that preventive measures can be implemented promptly.

In our problem setting, besides the topological structure, we also consider signals collected on the graph nodes, e.g., traffic density and traveling speed recorded at each sensor station in a traffic network.  The observed signals on one node over time form a time series.  We incorporate both the time series observations and evolving topological structure into our model for path classification.  The complex temporal dependency and structure dynamics pose a huge challenge.  For one thing, observations on nodes exhibit highly non-stationary properties such as seasonality or daily periodicity, e.g., morning and evening rush hours in a traffic network.  For another, graph structure evolution can result in sudden and dramatic changes of observations on nodes, e.g., road closure due to accidents redirects traffic to alternative routes, causing increased traffic flow on those routes.  To model the temporal dependency and structure dynamics, we design a new time-evolving neural network named Long Short-Term Memory R-GCN (LRGCN).  LRGCN considers node correlation within a graph snapshot as intra-time relation, and views temporal dependency between adjacent graph snapshots as inter-time relation, then utilizes Relational GCN (R-GCN)~\cite{schlichtkrull2018modeling} to capture both temporal dependency and structure dynamics.

Another challenge we face is that paths are of arbitrary length.  It is non-trivial to develop a uniform path representation that provides both good data interpretability and classification performance.  Existing solutions such as \cite{li2017deepcas} rely on Recurrent Neural Networks (RNN) to derive fixed-size representation, which, however, fails to provide meaningful interpretation of the learned path representation.  In this work, we design a new path representation method named \underline{s}elf-\underline{a}ttentive \underline{p}ath \underline{e}mbedding (SAPE), which takes advantage of the self-attentive mechanism to explicitly highlight the important nodes on a path, thus provides good interpretability and benefits downstream tasks such as path failure diagnosis.

Our contributions are summarized as follows.

\begin{itemize}
\item We study path classification in a time-evolving graph, which, to the best of our knowledge, has not been studied before.  Our proposed solution LRGCN achieves superior classification performance to the state-of-the-art deep learning methods.

\item We design a novel self-attentive path embedding method called SAPE to embed paths of arbitrary length into fixed-length vectors, which are then used as a standard input format for classification.  The embedding approach not only improves the classification performance, but also provides meaningful interpretation of the underlying data in two forms: (1) embedding vectors of paths, and (2) node importance in a path learned through a self-attentive mechanism that differentiates their contribution in classifying a path.

\item We evaluate LRGCN-SAPE on two real-world data sets.  In a telecommunication network of a real service session, we use LRGCN-SAPE to predict path failures and achieve a Macro-F1 score of 61.89\%, outperforming competing methods by at least 5\%.  In a traffic network in California, we utilize LRGCN-SAPE to predict path congestions and achieve a Macro-F1 score of 86.74\%, outperforming competing methods by at least 4\%.

\end{itemize}

The remainder of this paper is organized as follows.  Section \ref{def} gives the problem definition and Section \ref{alt} describes the design of LRGCN and SAPE.  We report the experimental results in Section \ref{sec.exp} and discuss related work in Section \ref{sec.related}.  Finally, Section \ref{sec.con} concludes the paper.

\begin{figure}
\begin{center}
\includegraphics [width=0.5\textwidth]{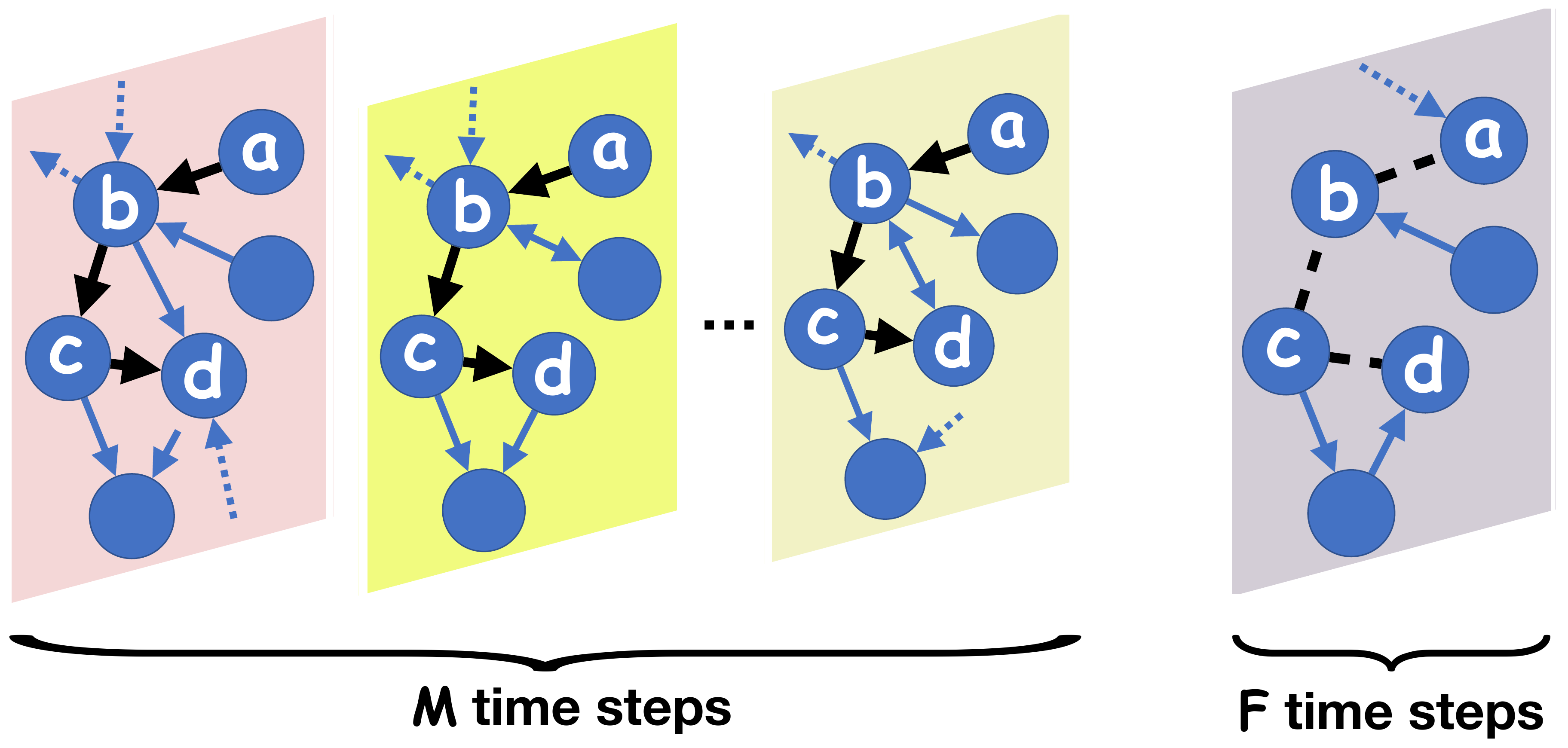}
\end{center}

\caption{A time-evolving graph in which four nodes $\{a,b,c,d\}$ correspond to four switches in a telecommunication network. Given the observations and graph snapshots in the past $M$ time steps, we want to infer if path $\left<a,b,c,d\right>$ will fail or not in the next $F$ time steps.}
\label{fig.example1}
\vspace{-0.3cm}
\end{figure}

\section{Problem Definition}\label{def}
We denote a set of nodes as $V=\{v_1, v_2, \ldots, v_N\}$ which represent real-world entities, e.g., switches in a telecommunication network, sensor stations in a traffic network.  At time $t$, we use an $N\times N$ adjacency matrix $A^t$ to describe the connections between nodes in $V$.  $A^t_{ij}\in \{0, 1\}$ represents whether there is a directed edge from node $v_j$ to node $v_i$ or not, e.g., a link that bridges two switches, a road that connects two sensor stations.  In this study, we focus on a directed graph, as many real-world networks, e.g., telecommunication networks, traffic networks, are directed; yet our methodology is also applicable to undirected graphs.  We use $X^t=\{x^t_1, x^t_2, \ldots, x^t_N\}$ to denote the observations at each node at time $t$, where $x^t_i\in \mathbb{R}^{d}$ is a $d$-dimensional vector describing the values of $d$ different signals recorded at node $v_i$ at time $t$, e.g., temperature, power and other signals of a switch.


We define the adjacency matrix $A^t$ and the observed signals $X^t$ on nodes in $V$ as a \emph{graph snapshot} at time $t$.  A sequence of graph snapshots with $\emph{\textbf{A}}=\{A^0, A^1, \ldots, A^t\}$ and the corresponding observations $\emph{\textbf{X}}=\{X^0, X^1, \ldots, X^t\}$ over time steps $0, 1, \ldots, t$ is defined as a \emph{time-evolving graph}.  Note that the graph structure can evolve over time as some edges may become unavailable, e.g., link failure, road congestion/closure, and some new edges may become available over time.  For one node $v_i\in V$, the sequence of observations $\{x^0_i, x^1_i, \ldots, x^t_i\}$ over time is a multivariate time series.

We denote a \emph{path} as a sequence $p = \left<v_1, v_2, \ldots, v_m \right>$ of length $m$ in the time-evolving graph, where each node $v_i \in V$\eat{ and there is a directed link between every two adjacent nodes in $p$}.  For the same path, we use $s^t = \left<x_1^t, x_2^t, \ldots, x_m^t \right>$ to represent the observations of the path nodes at time $t$.   In this paper we aim to predict if a given path is available or not in the future, e.g., a path failure in a telecommunication network, or a path congestion in a traffic network.  Note the \emph{availability} of a path is service dependent, e.g., a path is defined as \emph{available} in a telecommunication network if the transmission latency for a data packet to travel through the path is less than a predefined threshold.  Thus the path availability cannot be simply regarded as the physical connectivity of the path, but is related to the ``quality of service'' of the path.  To be more specific, for a given path $p$ at time $t$, we utilize the past $M$ time steps to predict the availability of this path in the next $F$ time steps.  We formulate this prediction task as a classification problem and our goal is to learn a function $f(\cdot)$ that can minimize the cross-entropy loss $\mathcal{L}$ over the training set $D$:
\begin{equation}
\arg\min \mathcal{L} = -\sum_{\emph{\textbf{P}}_j\in D}\sum_{c = 1}^CY_{jc} \log f_c(\emph{\textbf{P}}_j),
\label{equ.defi}
\end{equation}
where $\emph{\textbf{P}}_j = ([s_j^{t-M+1}, \ldots, s_j^t], p_j, [A^{t-M+1}, \ldots, A^t])$ is a training instance, $Y_j \in \{0, 1\}^C$ is the training label representing the availability of this path in the next $F$ time steps, $f_c(\emph{\textbf{P}}_j)$ is the predicted probability of class $c$, and $C$ is the number of classes.  In our problem, we have $C=2$, i.e., path availability and path failure.

Figure \ref{fig.example1} depicts a time-evolving graph in the context of a telecommunication network.  $a, b, c, d$ denote four switches.  In the past $M$ time steps, although the graph structure has changed, e.g., $\left<b, d\right>$ becomes unavailable due to overload, path $\left<a, b, c, d\right>$ is still available. From this time-evolving graph, we want to predict the availability of path $\left<a, b, c, d\right>$ in the next $F$ time steps.

\begin{figure*}
\begin{center}
\includegraphics [width=1\textwidth]{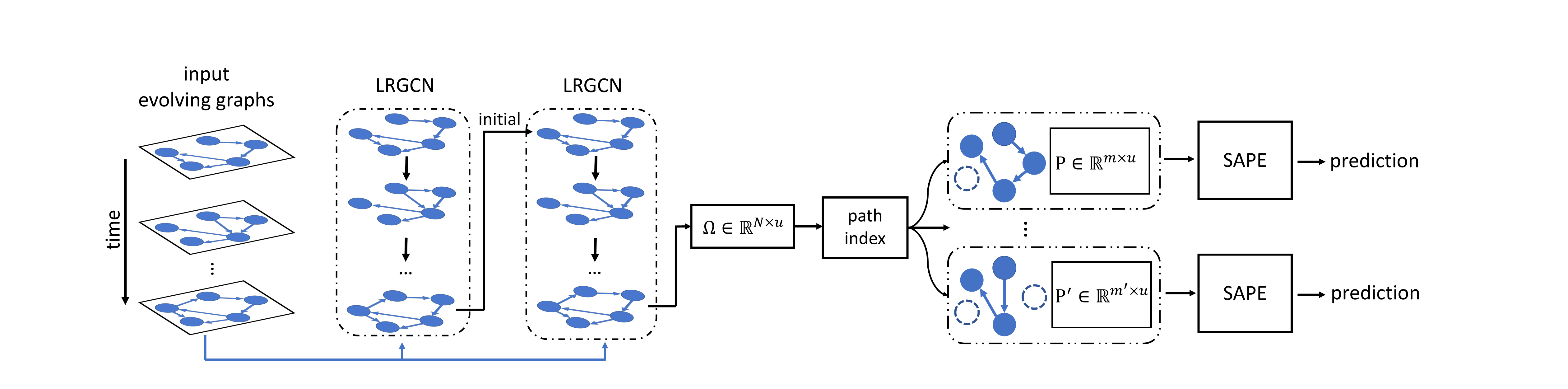}
\label{fig.em}
\end{center}

\caption{Schematic diagram of the framework designed for path classification.  The time-evolving graph in the past $M$ time steps is fed into a LRGCN layer whose final states are used to initialize another LRGCN layer. Paths are then indexed from the outputs of this LRGCN layer and SAPE is utilized to derive the final path representation for prediction.}
\label{fig.em}
\vspace{-0.3cm}
\end{figure*}

\section{Methodology}\label{alt}

\subsection{Framework}
In the context of time-evolving graph, we observe three important properties as follows.

\noindent\textbf{Property 1. Node correlation}.  Observations on nodes are correlated.  For example, if a sensor station in a traffic network detects low traffic density at time $t$, we can infer that nearby stations on the same path also record low traffic at the same time with high probability.

\noindent\textbf{Property 2. Influence of graph structure dynamics}.  Observations on nodes are influenced by the changes on the graph structure.  For example, if a road segment becomes unavailable at time $t$ (e.g., road closure), traffic shall be redirected to alternative routes.  As a result, nodes on the affected path may record a sudden drop of traffic density, while nodes on alternative routes may record an increase of traffic flow at the subsequent time steps which may increase the probability of path congestion.

\noindent\textbf{Property 3. Temporal dependency}.  The time series recorded on each node demonstrates strong temporal dependency, e.g., high traffic density and low traveling speed recorded at morning and evening rush hours.  This makes the time series non-stationary.

These three properties make our problem very complicated.  A desired model for path failure prediction should have built-in mechanisms to address these challenges.  First, it should model the node correlation and the influence of graph structure dynamics for an accurate prediction.  Second, it should capture the temporal dependency especially the long term trends from the time series.  Moreover, the temporal dependency and graph structure dynamics should be modeled jointly.  Third, the model should be able to represent paths of arbitrary length and generate a fixed-length representation by considering all path nodes as a whole.


In this section, we present a novel end-to-end neural network framework to address the above three requirements.  The framework (shown in Figure \ref{fig.em}) takes as input the time-evolving graph, and outputs the representation and failure probabilities for all the training paths.  To be more specific, our model uses a two-layer Long Short-Term Memory R-GCN (LRGCN), a newly proposed time-evolving graph neural network in this work, to obtain the hidden representation of each node by capturing both graph structure dynamics and temporal dependency.  Then it utilizes a self-attentive mechanism to learn the node importance and encode it into a unified path representation.  Finally, it cascades the path representation with a fully connected layer and calculates the loss defined in Eq. \ref{equ.defi}.  In the following, we describe the components in our model in details.

\subsection{Time-Evolving Graph Modeling}\label{demCOCC}
We propose a new time-evolving neural network to capture the graph structure dynamics and temporal dependency jointly.  Our design is mainly motivated by the recent success of graph convolutional networks (GCN) ~\cite{kipf2017semi} in graph-based learning tasks.  As GCN cannot take both time series $\emph{\textbf{X}}$ and evolving graph structures $\emph{\textbf{A}}$ as input, our focus is how to generalize GCN to process time series and evolving graph structures simultaneously.  In the following we first describe how we model the node correlation within a graph snapshot.  Then we detail temporal dependency modeling between two adjacent graph snapshots.  Finally, we generalize our model to the time-evolving graph.

\subsubsection{Static graph modeling}\label{dem}

Within one graph snapshot, the graph structure does not change, thus it can be regarded as a static graph.  The original GCN ~\cite{kipf2017semi} was designed to handle undirected static graphs.  Later, Relational GCN (R-GCN)~\cite{schlichtkrull2018modeling} was developed to deal with multi-relational graphs.  A directed graph can be regarded as a multi-relational graph with incoming and outgoing relations.  In this vein, we use R-GCN to model the node correlation in a static directed graph.  Formally, R-GCN takes as input the adjacency matrix $A^t$ and time series $X^t$, and transforms the nodes' features over the graph structure via one-hop normalization:
\begin{equation}
  Z = \sigma (\sum_{\phi \in R}{(D^{t}_{\phi})}^{-1}A^t_{\phi}X^tW_{\phi} + X^tW_0),
\label{equ.H}
\end{equation}
where $R = \{in, out\}$, $A^t_{in} = A^t$ represents the incoming relation, $A^t_{out} = (A^t)^T$ represents the outgoing relation and ${(D^t_{\phi})}_{ii} = \sum_j\ {(A^t_{\phi})}_{ij}$.  $\sigma(\cdot)$ is the activation function such as $ReLU(\cdot)$.  Eq. \ref{equ.H} can be considered as an accumulation of multi-relational normalization where $W_{in}$ is a weight matrix for incoming relation, $W_{out}$ for outgoing relation and $W_0$ for self-connection.

To further generalize R-GCN and prevent overfitting, we can view the effect of self-connection normalization as a linear combination of incoming and outgoing normalization. This provides us with the following simplified expression:
\begin{equation}
  Z_s = \sigma (\sum_{\phi \in R}\tilde{A}^t_{\phi}X^tW_{\phi}),
\label{equ.H2}
\end{equation}
where $\tilde{A}^t_{\phi} = (\hat{D}^t_{\phi})^{-1}\hat{A}^t_{\phi}$, $\hat{A}^t_{\phi} = {A^t_{\phi}} + I_N$, $(\hat{D}^t_{\phi})_{ii} = \sum_j\ (\hat{A}^t_{\phi})_{ij}$, and $I_N$ is the identity matrix.  Note we can impose multi-hop normalization by stacking multiple layers of R-GCN.  In our design, we use a two-layer R-GCN:
\begin{equation}
  \Theta_s {\star}g\ X^t = \sum_{\phi \in R}\tilde{A}^t_{\phi}\sigma (\sum_{\phi \in R}\tilde{A}^t_{\phi}X^tW_{\phi}^{(0)})W^{(1)}_{\phi},
\label{equ.H3}
\end{equation}
where $\Theta_s$ represents the parameter set used in the static graph modeling, $W_{\phi}^{(0)} \in \mathbb{R}^{d \times h}$ is an input-to-hidden weight matrix for a hidden layer with $h$ feature maps.  $W_{\phi}^{(1)} \in \mathbb{R}^{h \times u}$ is a hidden-to-output weight matrix, ${\star}g$ stands for this two-hop graph convolution operation and shall be used thereafter.

\noindent{\textbf{Relation with the original GCN}}.  The original GCN ~\cite{kipf2017semi} was defined on undirected graphs and can be regarded as a special case of this revised R-GCN.  One difference is that in undirected graphs incoming and outgoing relations are identical, which makes $W_{in} = W_{out}$ in R-GCN for undirected graphs.  Another difference lies in the normalization trick.  The purpose of this trick is to normalize features of each node according to its one-hop neighborhood.  In undirected graphs the relation is symmetric, thus the symmetric normalization is applied as $D^{-\frac{1}{2}}AD^{-\frac{1}{2}}$, where $D_{ii} = \sum_j\ A_{ij}$; while in directed graphs the relation is asymmetric, hence the asymmetric normalization $D^{-1}A$ is used.

The discussion above focuses on a graph snapshot, which is static.  Next, we extend R-GCN to take as inputs two adjacent graph snapshots.

\subsubsection{Adjacent graph snapshots modeling}\label{ssc}
\begin{figure}
\begin{center}
\includegraphics [width=0.4\textwidth]{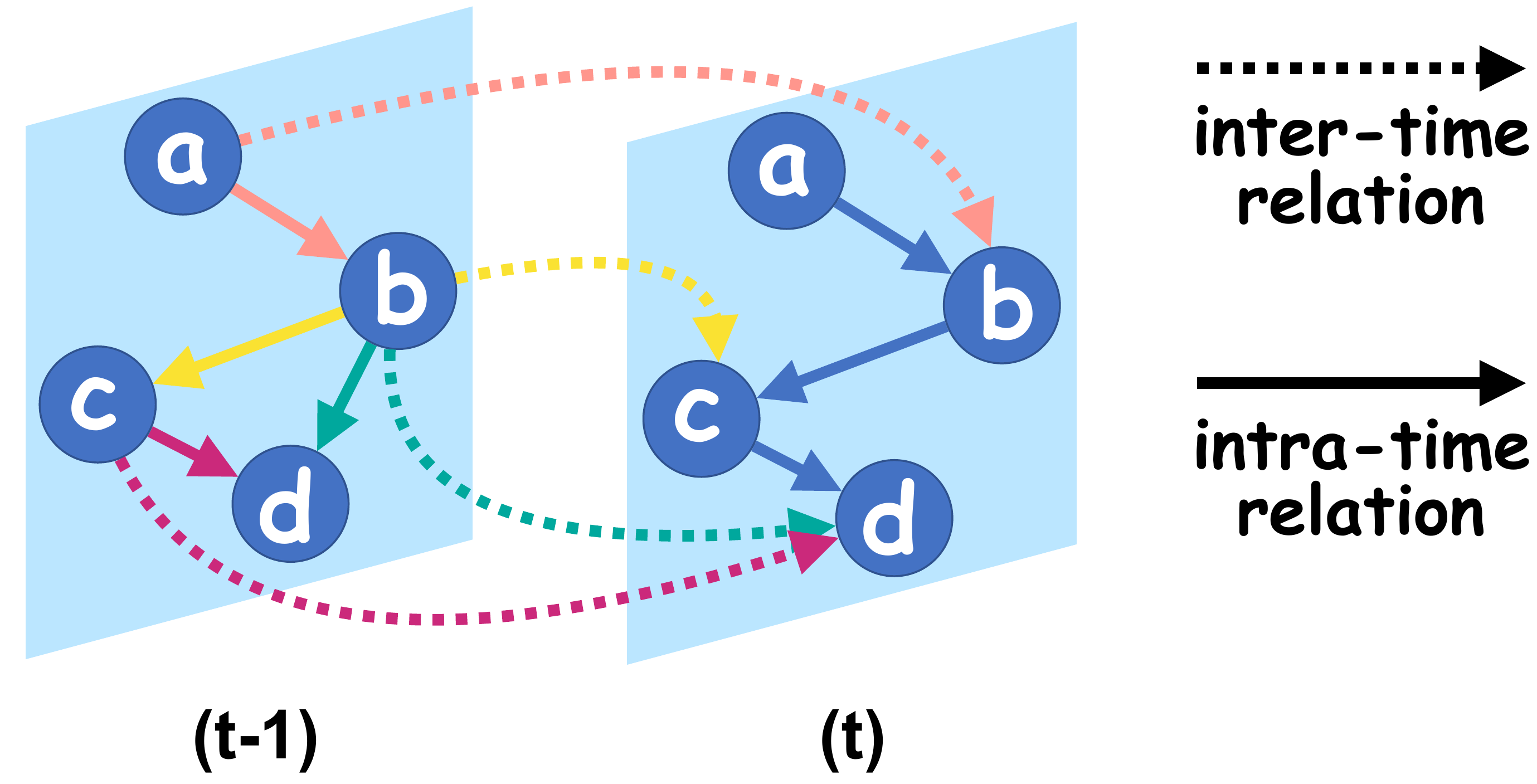}
\end{center}

\caption{Plot of intra-time relation (in solid line) and inter-time relation (in dotted line) modeled for two adjacent graph snapshots.}
\label{fig.prop}
\eat{\vspace{-0.3cm}}
\end{figure}

Before diving into a sequence of graph snapshots, we first focus on two adjacent time steps $t-1$ and $t$ as shown in Figure \ref{fig.prop}.  A node at time $t$ is not only correlated with other nodes at the same time (which is referred to as \emph{intra-time relation}), but also depends on nodes at the previous time step $t-1$ (which is referred to as \emph{inter-time relation}), and this dependency is directed and asymmetric.  For example, if a sensor station detects high traffic density at time $t$, then nearby sensor stations may also record high traffic density at the same time due to the spatial proximity.  Moreover, if a sensor station detects a sudden increase of traffic density at time $t-1$, downstream stations on the same path will record the corresponding increase at subsequent time steps, as it takes time for traffic flows to reach downstream stations.  In our model, we use the Markov property to model the inter-time dependency.  In total, there are four types of relations to model in R-GCN, i.e., intra-incoming, intra-outgoing, inter-incoming and inter-outgoing relations.  For nodes at time $t$, the multi-relational normalization expression is as follows:
\begin{equation}
  G\_unit(\Theta,[X^t,X^{t-1}]) = \sigma (\Theta_s {\star}g\ X^t + \Theta_h {\star}g\ X^{t-1}),
\label{equ.H4}
\end{equation}
where $\Theta_h$ stands for the parameter set used in inter-time modeling, and it does not change over time\eat{, i.e., stationary}.  For $\Theta_h {\star}g\ X^{t-1}$, $\tilde{A}^{t-1}_{\phi}$ is used to represent the graph structure.  This operation is named time-evolving graph \textbf{G\_unit}, which has a similar role of unit in Recurrent Neural Networks (RNN).  Note that here the normalization still includes inter-time self-connection, as $\tilde{A}^{t-1}_{\phi}$ has self-loops.

Intuitively, Eq. \ref{equ.H4} computes the new feature of a node by accumulating transformed features via a normalized sum of itself and neighbors from both the current and previous graph snapshots.  As nodes which are densely connected by inter-time and intra-time relations tend to be proximal, this computation makes their representation similar, thus simplifies the downstream tasks.

\noindent{\textbf{Relation with RNN unit}}.  RNN unit was proposed to transform an input by considering not only the present input but also the input preceding it in a sequence.  It can be regarded as a special case of our time-evolving graph unit where at each time step a set of input elements are not structured and $\tilde{A}_{\phi} = I_N$ if we consider just one-hop smoothing.

\begin{figure*}
\begin{center}
\includegraphics [width=0.7\textwidth]{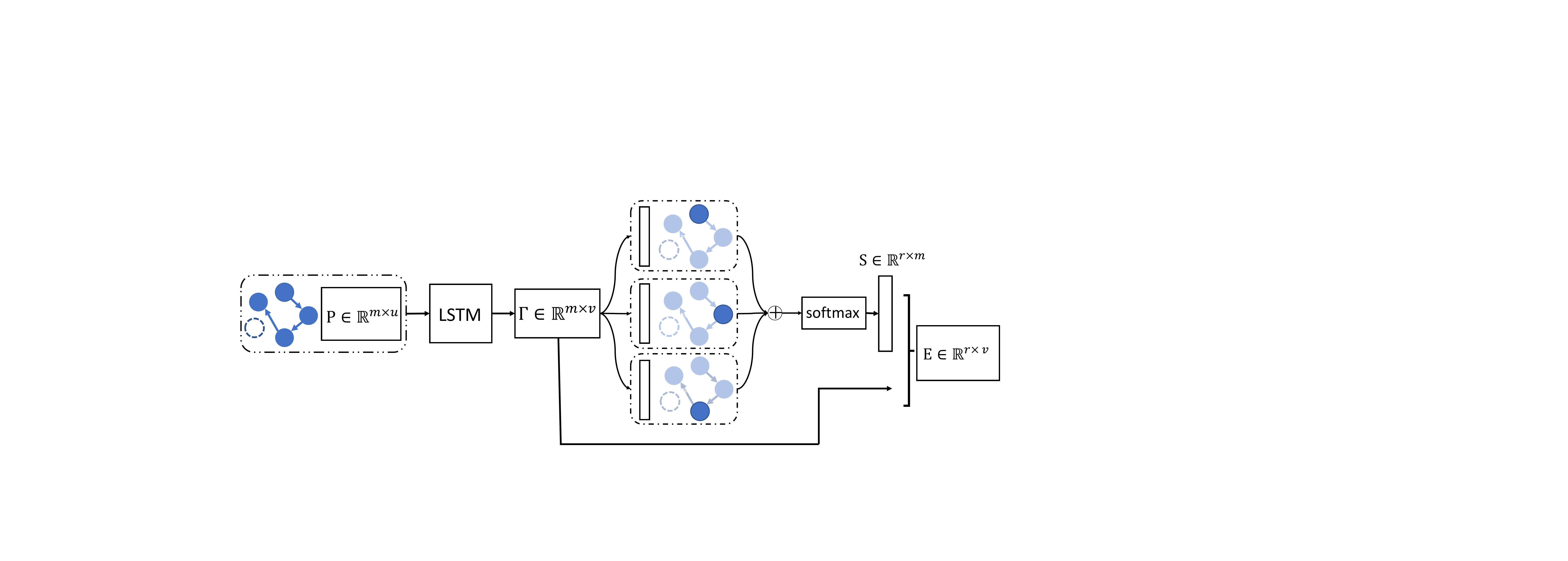}
\end{center}

\caption{The proposed self-attentive path embedding method SAPE.}
\label{fig.sape}
\eat{\vspace{-0.3cm}}
\end{figure*}

\subsubsection{The proposed LRGCN model}\label{lrgcn}
Based on the time-evolving graph unit proposed above, we are ready to design a neural network working on a time-evolving graph.  We use a hidden state $H^{t-1}$ to memorize the transformed features in the previous snapshots, and feed the hidden state and current input into the unit to derive a new hidden state:
\begin{equation}
  H^{t} = \sigma (\Theta_H {\star}g\ [X^t,H^{t-1}]),
\label{equ.HH}
\end{equation}
where $\Theta_H$ includes $\Theta_s$ and $\Theta_h$. When applied sequentially, as the transformed features in $H^{t-1}$ can contain information from an earlier arbitrarily long window, it can be utilized to process a sequence of graph snapshots, i.e., a time-evolving graph.

Unfortunately, despite the usefulness of this RNN-style evolving graph neural network, it still suffers from the famous curse of gradient exploding or vanishing.  In this context, past studies (e.g., \cite{sutskever2014sequence}) utilized Long Short-Term Memory (LSTM) \cite{hochreiter1997long} to model the long-term dependency in sequence learning.  Inspired by this, we propose a Long Short-Term Memory R-GCN, called LRGCN, which can take as input a long-term time-evolving graph and capture the structure dynamics and temporal dependency jointly.  Formally, LRGCN utilizes three gates to achieve the long-term memory or accumulation:
\begin{equation}
  \textbf{i}^t = \sigma (\Theta_i {\star}g\ [X^t,H^{t-1}])
\label{equ.H5}
\end{equation}
\begin{equation}
  \textbf{f}^t = \sigma (\Theta_f {\star}g\ [X^t,H^{t-1}])
\label{equ.H6}
\end{equation}
\begin{equation}
  \textbf{o}^t = \sigma (\Theta_o {\star}g\ [X^t,H^{t-1}])
\label{equ.H7}
\end{equation}
\begin{equation}
  \textbf{c}^t = \textbf{f}^t \odot\ \textbf{c}^{t-1} + \textbf{i}^t \odot \tanh (\Theta_c {\star}g\ [X^t,H^{t-1}])
\label{equ.H8}
\end{equation}
\begin{equation}
  H^t = \textbf{o}^t \odot \textbf{c}^t
\label{equ.H9}
\end{equation}
where $\odot$ stands for element-wise multiplication, $\textbf{i}^t$, $\textbf{f}^t$, $\textbf{o}^t$ are input gate, forget gate and output gate at time $t$ respectively.  $\Theta_i$, $\Theta_f$, $\Theta_o$, $\Theta_c$ are the parameter sets for the corresponding gates and cell.  $H^t$ is the hidden state or output at time $t$, as used in Eq. \ref{equ.HH}.  ${\star}g$ denotes the two-hop graph convolution operation defined in Eq. \ref{equ.H3}.  Intuitively, LRGCN achieves long-term time-evolving graph memory by carefully selecting the input to alter the state of the memory cell and to remember or forget its previous state, according to the tasks at hand.

To summarize, we use a two-layer LRGCN to generate hidden representation of each node, in which the first-layer LRGCN serves as the encoder of the whole time-evolving graph and is used to initialize the second-layer LRGCN.  Then we take the outputs of the last time step in the second-layer LRGCN to derive the final representation $\Omega \in \mathbb{R}^{N \times u}$.  As we work on path classification, the next task is how to obtain the path representation based on $\Omega$.

\subsection{Self-Attentive Path Embedding}\label{SAPE}

In this subsection, we describe our method which produces a fixed-length path representation given the output of the previous subsection.  For a path instance $\emph{\textbf{P}}$, we can retrieve its representation $P \in \mathbb{R}^{m \times u}$ directly from $\Omega$.  For the final classification task, however, we still identify two challenges:

\begin{itemize}
\item  \emph{Size invariance}: How to produce a fixed-length vector representation for any path of arbitrary length?
\item  \emph{Node importance}: How to encode the importance of different nodes into a unified path representation?
\end{itemize}

For \emph{node importance}, it means different nodes in a path have different degrees of importance.  For example, along a path, a sensor station at the intersection of main streets should be more important than one at a less busy street in contributing to the derived embedding vector.  We need to design a mechanism to learn the node importance, and then encode it in the embedding vector properly.

To this end, we propose a self-attentive path embedding method, called SAPE, to address the challenges listed above.  In SAPE, we first utilize LSTM to sequentially take in node representation of a path, and output representation in each step by balancing upstreaming node representation and current input node representation, as proposed in ~\cite{li2017deepcas,LiuZZZCWY17}.  Then we use the self-attentive mechanism to learn the node importance and transform a path of variable length into a fixed-length embedding vector.  Figure~\ref{fig.sape} depicts the overall framework of SAPE.

Formally, for a path $P \in \mathbb{R}^{m \times u}$, we first apply LSTM to capture node dependency along the path sequence:
\begin{equation}
  \Gamma = \textsf{LSTM}(P),
\label{equ.gene}
\end{equation}
where $\Gamma \in \mathbb{R}^{m \times v}$. With $\textsf{LSTM}$, we have transformed the node representation from a $u$-dimensional space to a $v$-dimensional space by capturing node dependency along the path sequence.

Note that this intermediate path representation $\Gamma$ does not provide node importance, and it is size variant, i.e., its size is still determined by the number of nodes $m$ in the path.  So next we utilize the self-attentive mechanism to learn node importance and encode it into a unified path representation, which is size invariant:
\begin{equation}
  S = \textsf{softmax} \big(W_{h2}\textsf{tanh}(W_{h1}\Gamma^T)\big),
\label{equ.gene}
\end{equation}
where $W_{h1} \in \mathbb{R}^{d_s \times v}$ and $W_{h2} \in \mathbb{R}^{r \times d_s}$ are two weight matrices. The function of $W_{h1}$ is to transform the node representation from a $v$-dimensional space to a $d_s$-dimensional space.  $W_{h2}$ is used as $r$ views of inferring the importance of each node.  Then \textsf{softmax} is applied to derive a standardized importance of each node, which means in each view the summation of all node importance is 1.

Based on all the above, we compute the final path representation $E \in \mathbb{R}^{r \times v}$ by multiplying $S \in \mathbb{R}^{r \times m}$ with $\Gamma \in \mathbb{R}^{m \times v}$:
\begin{equation}
E = S\Gamma.
\end{equation}
$E$ is size invariant since it does not depend on the number of nodes $m$ any more.  It also unifies the node importance into the final representation, in which the node importance is only determined by the downstream tasks.

\eat{
One potential risk in SAPE is that $r_s$ or $r_h$ views may be similar. To diversify their views, a penalization term is imposed:
\begin{equation}
  P = \big|\big|\ S_T(S_T)^T - I_{r_s}\ \big|\big|_F^2 + \big|\big|\ SS^T - I_{r_h}\ \big|\big|_F^2.
\end{equation}
Here $\big|\big|\cdot\big|\big|_F$ represents the Frobenius norm of a matrix.  We train the classifier in a supervised way with the task at hand, in the hope of minimizing both the penalization and the cross-entropy loss.
}

In all, our framework first uses a two-layer LRGCN to obtain hidden representation of each node by capturing graph structure dynamics and temporal dependency.  Then it uses SAPE to derive the path representation that takes node importance into consideration. The output of SAPE is cascaded with a fully connected layer to compute the final loss.

\section{EXPERIMENTS}\label{sec.exp}
We validate the effectiveness of our model \eat{through the prediction tasks }on two real-world data sets: (1) predicting path failure in a telecommunication network, and (2) predicting path congestion in a traffic network.

\subsection{Data}
\subsubsection{Telecommunication network (Telecom)}
This data set targets a metropolitan LTE transmission network serving 10 million subscribers.  We select 626 switches and collect 3 months of data from Feb 1, 2018 to Apr 30, 2018.  For each switch, we collect two values every 15 minutes: sending power and receiving power. From the network, we construct an adjacency matrix $A$ by denoting $A_{ij} = 1$ if there is a directed optical link from $v_j$ to $v_i$, and $A_{ij} = 0$ otherwise.  The graph structure changes when links fail or recover.  The observations on switches and the graph structure over time form a time-evolving graph, where a time step corresponds to 15 minutes and the total number of time steps is 8449 over 3 months.
\begin{table}
  \caption{Statistics of path instances}
  \label{tab:twodatasets}
  \begin{tabular}{ccc}
    \toprule
&\textbf{Telecom}& \textbf{Traffic}\\
    \midrule
	No. of failure/congestion &385,896&85,083\\
	No. of availability &6,821,101&346,917\\
	Average length of paths &7.05\textpm 4.39&32.56\textpm 12.48\\
  \bottomrule
\end{tabular}
\vspace{-0.4cm}
\end{table}
There are 853 paths serving transmission of various services. \eat{ We label a path as failure if alarms are triggered on the path by the alarm system.  We use 24 hours' history data to predict if a path will fail in the next 24 hours, i.e., $M=96$ and $F=96$.  Under this setting, we have $8449\times 853$ samples, of which $2000\times 853$ samples are used for testing, $1000\times 853$ for validation, and $5449\times 853$ for training.}Using a sliding window over time, we create $8449 \times 853$ path instances, of which $5449 \times 853$ instances are used for training, $1000 \times 853$ for validation, and $2000 \times 853$ for testing.  We label a path instance at time $t$ as failure if alarms are triggered on the path by the alarm system within time steps $[t+1, t+F]$.  We use 24 hours' history data to predict if a path will fail in the next 24 hours, i.e., $M=96$ and $F=96$.
\subsubsection{Traffic network (Traffic)}
This data set targets District 7 of California collected from Caltrans Performance Measurement System (PeMS).  We select 4438 sensor stations and collect 3 months of data from Jun 1, 2018 to Aug 30, 2018.  For each station, we collect two measures: average speed and average occupancy at the hourly granularity by aggregation.  From the traffic network, we construct an adjacency matrix $A$ by denoting $A_{ij} = 1$ if $v_j$ and $v_i$ are adjacent stations on a freeway along the same direction.  The graph structure changes according to the node status (e.g., congestion or closure).  A time-evolving graph is constructed from observations recorded in stations and the dynamic graph structure, where a time step is an hour and the total number of time steps is 2160 over 3 months.

We sample 200 paths by randomly choosing two stations as the source and target, then use Dijkstra's algorithm to generate the shortest path.  \eat{We label a path as congestion if two consecutive stations are labeled as congestion.  We use 24 hours' history data to predict if a path will congest in the next one hour, i.e., $M = 24$ and $F = 1$.  Under this setting, we have $2160 \times 200$ samples, of which $500 \times 200$ are used for testing, $200 \times 200$ for validation, and $1460 \times 200$ for training.} Using a sliding window over time, we create $2160\times 200$ path instances, of which $1460\times 200$ instances are used for training, $200\times 200$ for validation, and $500\times 200$ for testing.  We label a path instance at time $t$ as congestion if two consecutive stations are labeled as congestion within time steps $[t+1, t+F]$.  We use 24 hours' history data to predict if a path will congest in the next one hour, i.e., $M=24$ and $F=1$.

\begin{figure}
\begin{center}
\includegraphics [width=0.37\textwidth,scale=1]{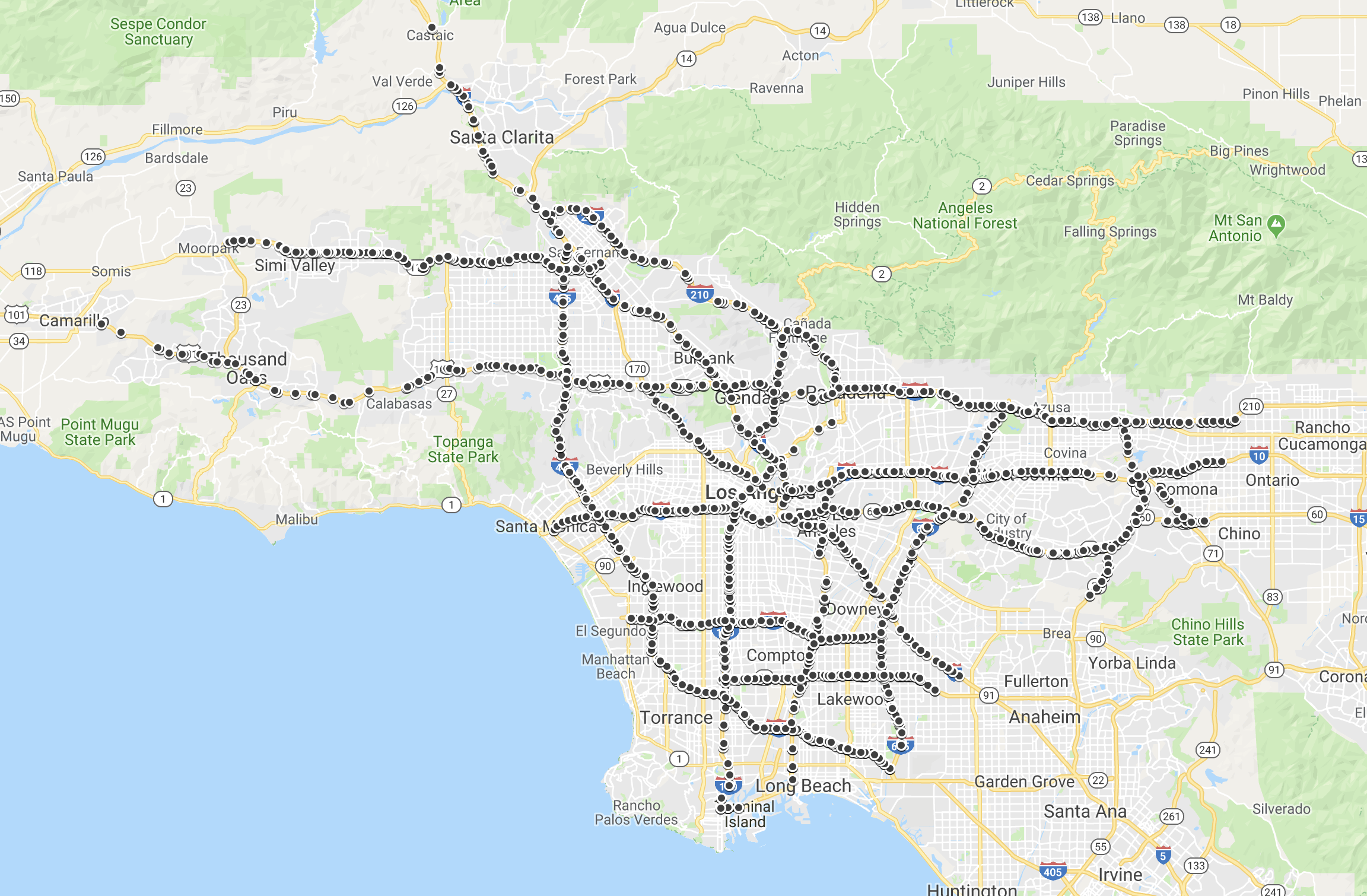}
\end{center}

\caption{Sensor distribution in District 7 of California. Each dot represents a sensor station.}
\label{fig.d7}
\vspace{-0.4cm}
\end{figure}

Table \ref{tab:twodatasets} lists the statistics of path instances in the two data sets.  Figure \ref{fig.d7} depicts the geographical distribution of sensors in Traffic data. Please refer to Appendix \ref{a.c} for detailed data preprocessing of the two data sets.

\subsection{Baselines and Metrics}\label{syn.base}
\eat{We use the following approaches as our baselines:}
\begin{itemize}
\item DTW \cite{dtw}, which first measures node similarity by the Dynamic Time Warping distance of time series observed on each node, then models the path as a bag of nodes, and calculates the similarity between two paths by their maximum node similarity.


\item FC-LSTM, which uses two-layer LSTM to capture the temporal dependency and uses another LSTM layer to derive the path representation. It only considers the time series sequences, but does not model node correlation or graph structure.

\item DCRNN \cite{li2018diffusion}, which uses two-layer DCRNN to capture both temporal dependency and node correlation and uses LSTM to get path representation from the last hidden state of the second DCRNN layer. It works on a static graph.

\item STGCN \cite{yu2018spatio}, which is similar to DCRNN except that we replace DCRNN with STGCN.

\item LRGCN, which is similar to DCRNN except that we replace DCRNN with LRGCN.

\item LRGCN-SAPE (static), which is similar to LRGCN except that we replace the path representation method LSTM with SAPE.

\item LRGCN-SAPE (evolving), which is similar to LRGCN-SAPE (static) except that the underlying graph structure evolves over time.
\end{itemize}

All neural network approaches are implemented using Tensorflow, and trained using minibatch based Adam optimizer with exponential decay.  The best hyperparameters are chosen using early stopping with an epoch window size of 3 on the validation set.  All trainable variables are initialized by He-normal \cite{he2015delving}. For fair comparison, DCRNN, STGCN and LRGCN use the same static graph structure and LSTM path representation methods.  Detailed parameter settings for all methods are available in Appendix \ref{a.b}.

We run each method three times and report the average Precision, Recall and Macro-F1.  Precision and Recall are computed with respect to the positive class, i.e., path failure/congestion, and defined as follows.

\begin{equation}
Precision = \frac{\text{\#true positives}}{\text{\#true positives} + \text{\#false positives}},
\end{equation}

\begin{equation}
Recall = \frac{\text{\#true positives}}{\text{\#true positives} + \text{\#false negatives}}.
\end{equation}
F1 score of the positive class is defined as follows:
\begin{equation}
F_1(P) = \frac{2 \times Precision \times Recall}{Precision + Recall}.
\end{equation}
Macro-F1 score is the average of F1 scores of the positive and negative classes, i.e., $F_1(P)$ and $F_1(N)$:
\begin{equation}
Macro{\text -}F1 = \frac{F_1(P) + F_1(N)}{2}.
\end{equation}

\eat{

 In our problem, Precision stands for the ratio of correct positive predictions to all the positive predictions (i.e., positive refers to path failure/congestion), and Recall measures the ratio of positive predictions to all the actual positives.  For detailed definitions of these metrics, please refer to Appendix \ref{a.d}.

Supporse $PP$ is the number of cases that are predicted as positive (here positive refers to path failure or path congestion), $AP$ is the number of cases that are actual positive, $TP$ is the number of cases that are correctly predicted as positive.
\begin{equation}
Precision = \frac{TP}{PP}
\end{equation}

\begin{equation}
Recall = \frac{TP}{AP}
\end{equation}

}

\subsection{Results}
\subsubsection{Classification performance}
Tables \ref{tele} and \ref{traffic} list the experimental results on Telecom and Traffic data sets respectively.  Among all approaches, LRGCN-SAPE (evolving) achieves the best performance.  In the following, we analyze the performance of all methods categorized into 4 groups.

\noindent\underline{Group 1}:  DTW performs worse than all neural network based methods.  One possible explanation is that DTW is an unsupervised method, which fails to generate discriminative features for classification.  Another possibility is that DTW measures the similarity of two time series by their pairwise distance and does not capture temporal dependency like its competitors.

\noindent\underline{Group 2}: FC-LSTM performs worse than the three neural network methods in Group 3, in both Macro-F1 and Precision, which proves the effectiveness of node correlation modeling in Group 3.

\noindent\underline{Group 3}: All the three neural networks in this group model both node correlation and temporal dependency, but the underlying graph structure is static and does not change.  LRGCN outperforms both DCRNN and STGCN by at least 1\% in Macro-F1 on both data sets, indicating LRGCN is more effective in node correlation and temporal dependency modeling.  For STGCN and DCRNN, DCRNN performs slightly better (0.19\% in Macro-F1) on Traffic data and STGCN performs better (1.87\% in Macro-F1) on Telecom data.

\noindent\underline{Group 4}: LRGCN-SAPE (static) works on a static graph and LRGCN-SAPE (evolving) works on a time-evolving graph.  LRGCN-SAPE (static) outperforms Group 3 methods by at least 1\% in Macro-F1 on both data sets, which means that SAPE is superior to pure LSTM in path representation.  LRGCN-SAPE (evolving) further achieves substantial improvement based on the time-evolving graph, i.e., it improves Macro-F1 by 1.34\% on Telecom and 1.90\% on Traffic.

\subsubsection{Training efficiency}
\begin{figure}
\begin{center}
\includegraphics [width=0.35\textwidth,scale=1]{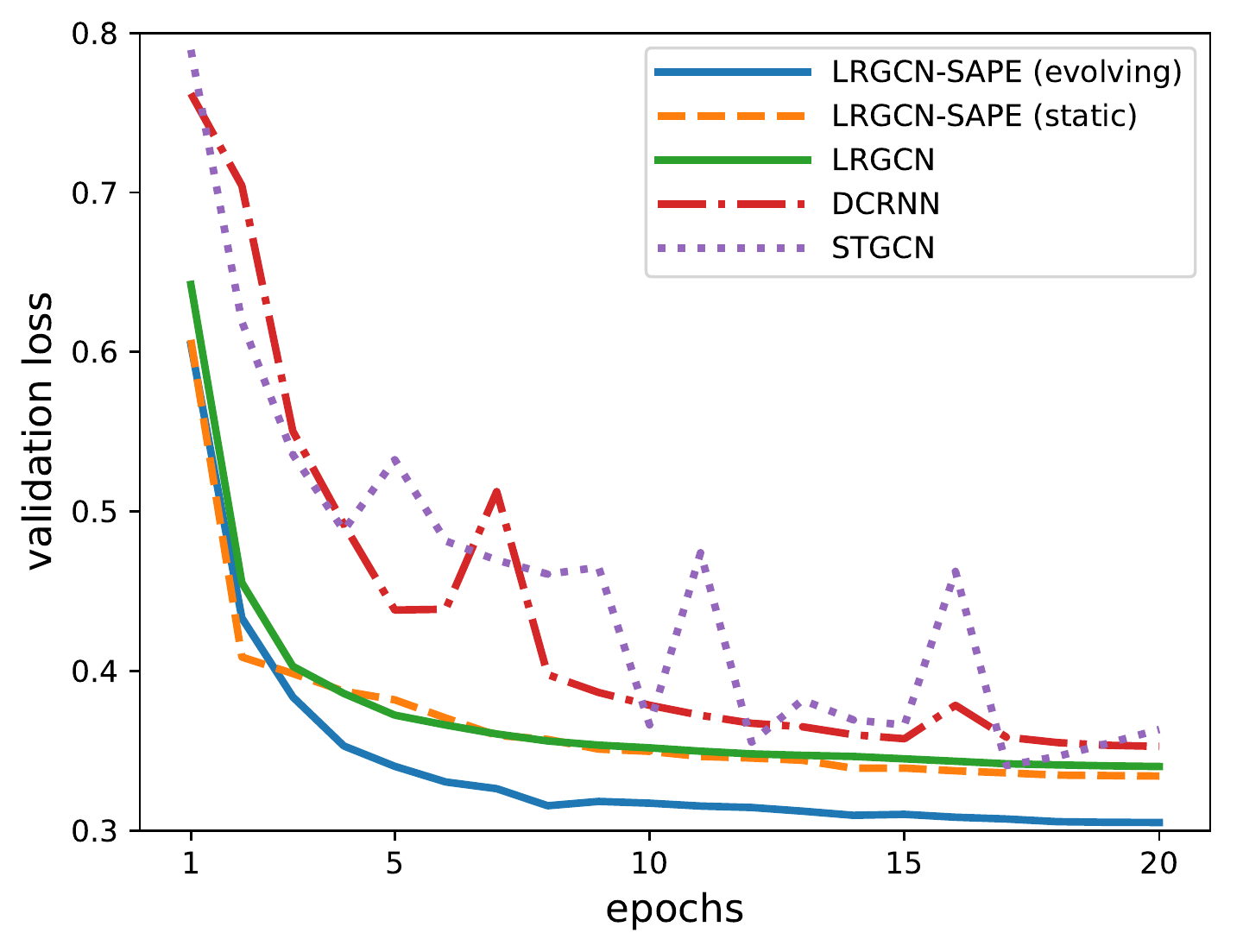}
\end{center}

\caption{Learning curve of different methods. LRGCN-SAPE (evolving) achieves the lowest validation loss.}
\label{fig.lcurve}
\end{figure}

To compare the training efficiency of different methods, we plot the learning curve of different methods in Figure \ref{fig.lcurve}.  We find that our proposed LRGCN-based methods including LRGCN, LRGCN-SAPE (static) and LRGCN-SAPE (evolving) converge more quickly than other methods.  Another finding is that after three epochs, LRGCN-SAPE (evolving) outperforms other methods by achieving the lowest validation loss, which indicates a better training efficiency of our proposed method on time-evolving graphs.

\subsubsection{Benefits of graph evolution modeling}
\begin{figure*}
\centering
\includegraphics [width=0.21\textwidth,height = 0.2\textheight]{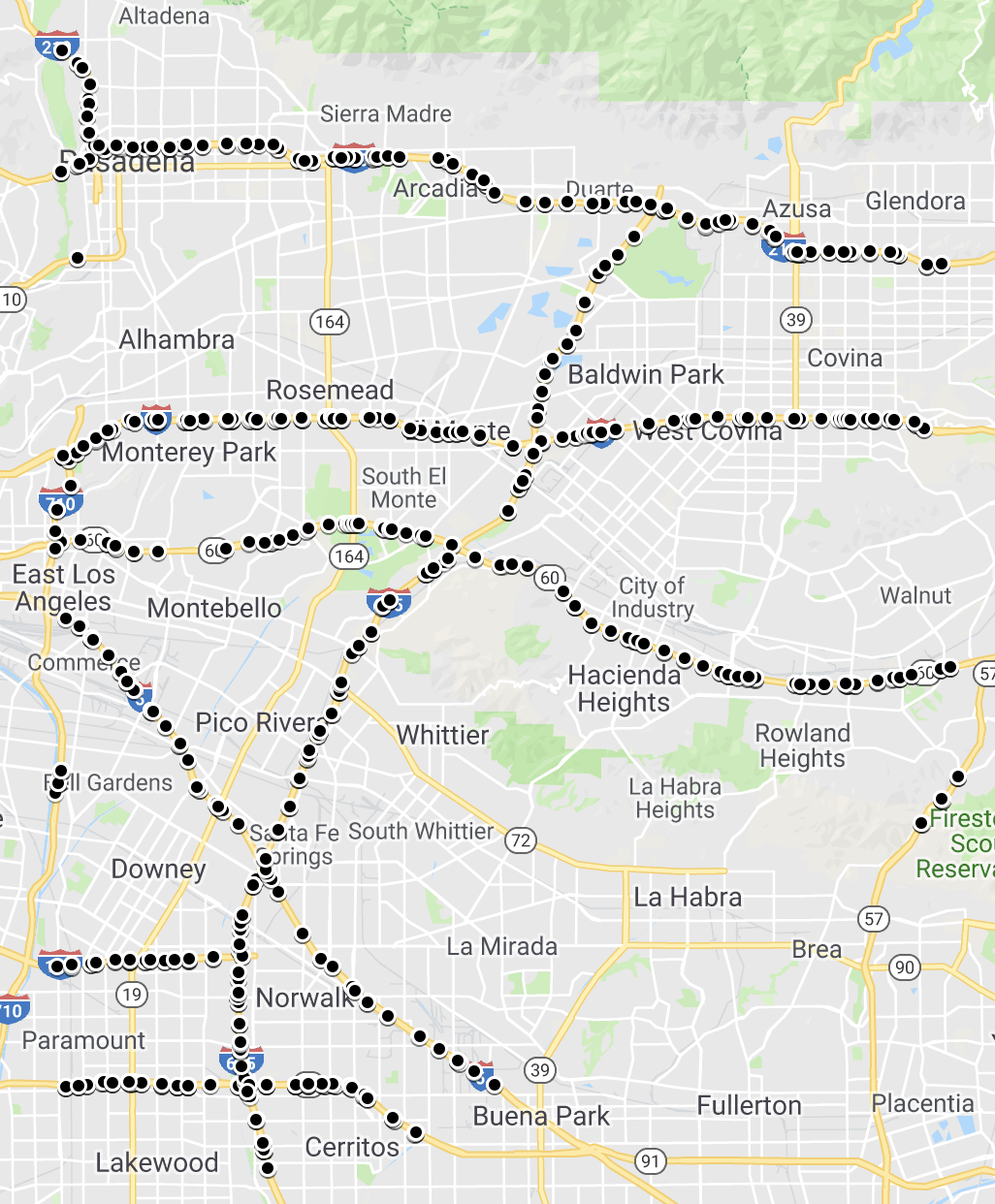}
\includegraphics [width=0.25\textwidth]{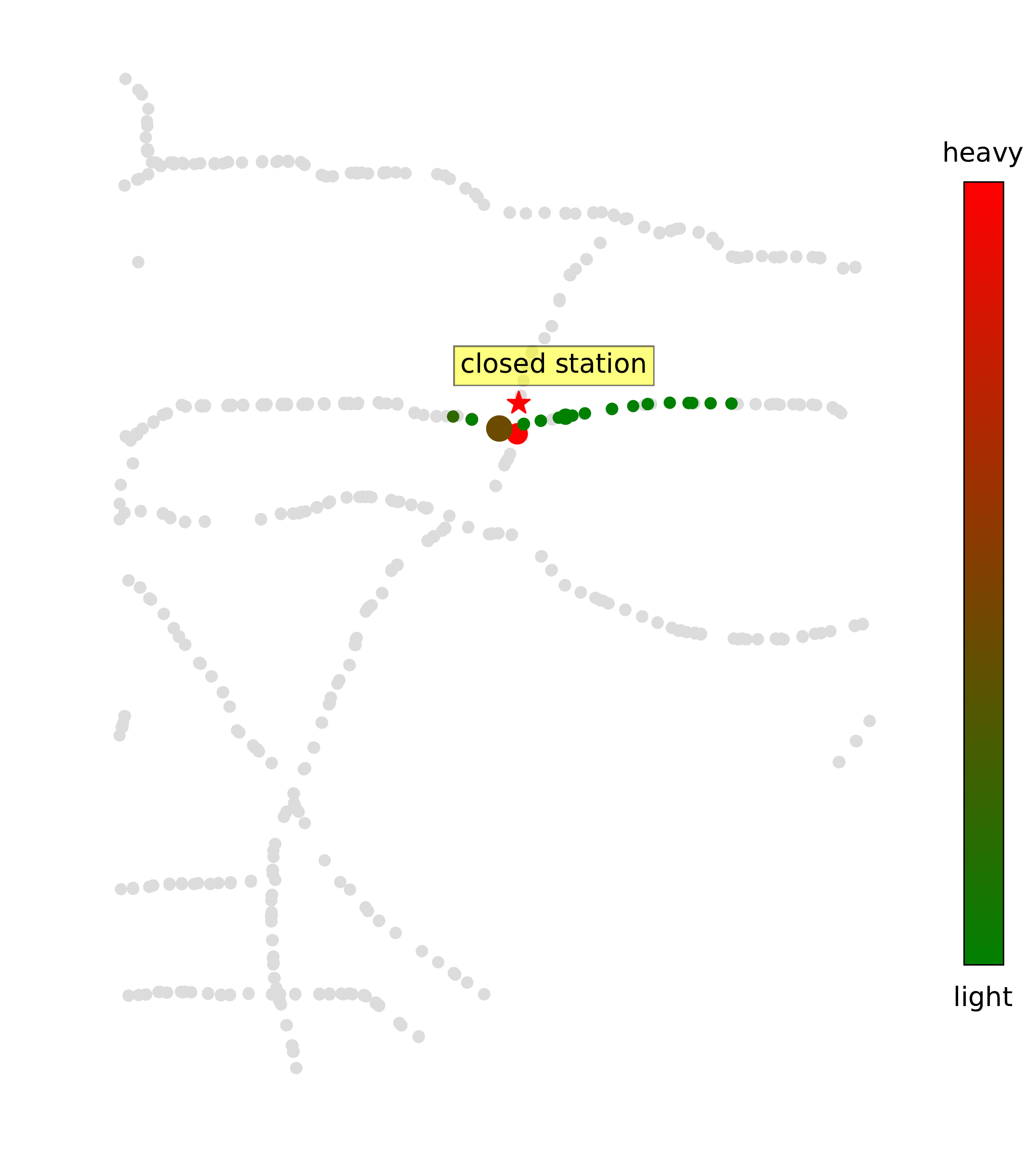}
\includegraphics [width=0.25\textwidth]{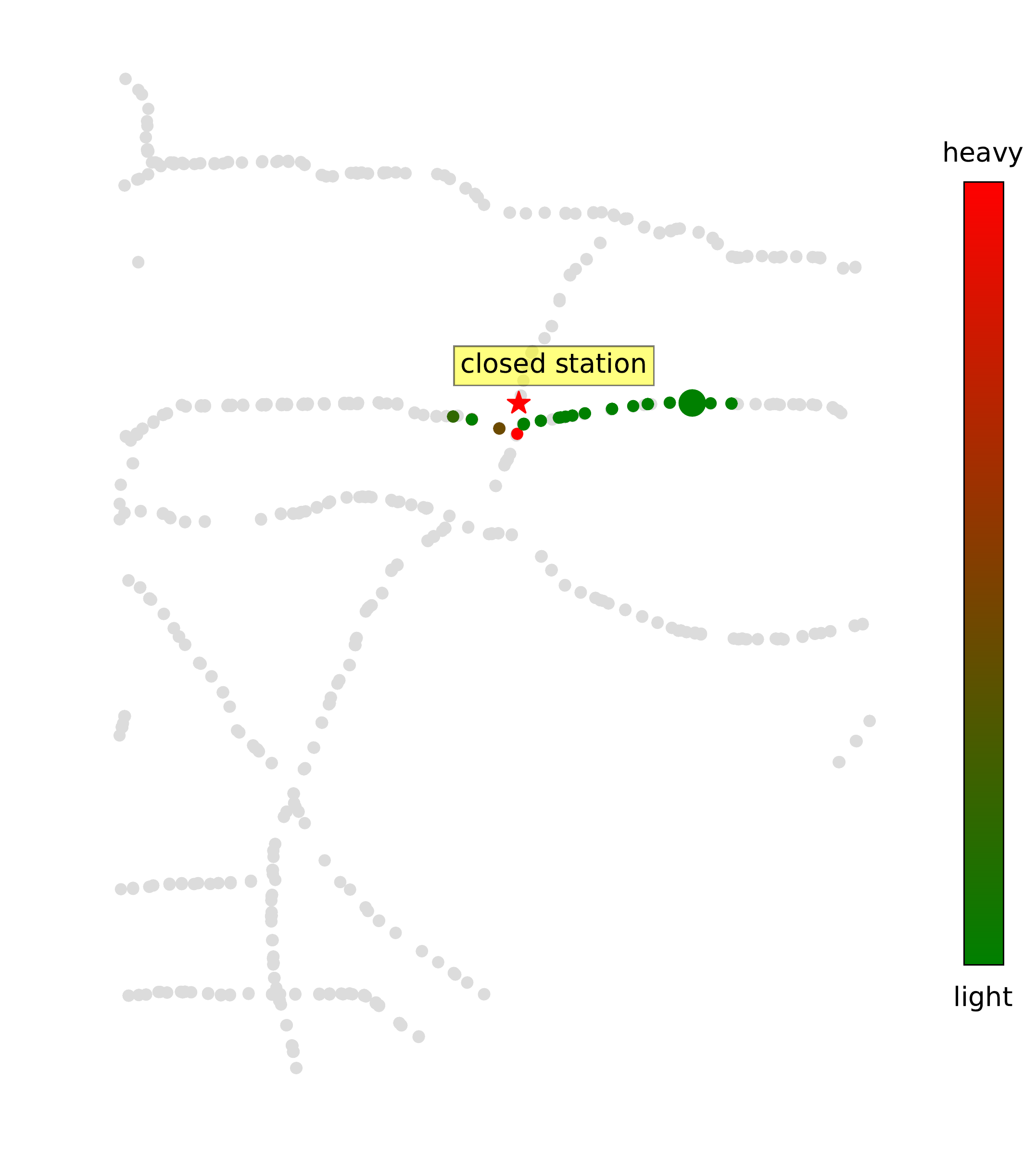}
\label{rhythm}
\caption{Visualization of learned attention weights of a path on Traffic (left: the original map; middle: attention weights by LRGCN-SAPE (evolving); right: attention weights by LRGCN-SAPE (static)). The star denotes a closed station. A bigger node radius indicates a larger attention weight. Red color represents heavy traffic recorded by sensor stations while green color represents light traffic.}
\label{fig.rhythm}
\end{figure*}

To further investigate how LRGCN performs on time-evolving graphs, we target a path which is heavily influenced by a closed sensor station and visualize the node attention weight learned by LRGCN-SAPE (evolving) and LRGCN-SAPE (static) respectively in Figure \ref{fig.rhythm}.  In the visualization, green color represents light traffic recorded by sensor stations, red color represents heavy traffic, and a bigger radius of a node denotes a larger attention weight, as the average across $r$ views inferred by SAPE.  We find that LRGCN-SAPE (evolving) can capture the dynamics of graph structure caused by the closure of the station, in the sense that the nearby station is affected subsequently, thus receives more attention.  In contrast, LRGCN-SAPE (static) is unaware of the graph structure change and assigns a large attention weight to a node that is far away from the closed station.

\begin{table}
  \caption{Comparison of different methods on path failure prediction on Telecom}
  \label{tele}
  \scalebox{0.9}{
  \begin{tabular}{ccccc}
    \toprule
  &\textbf{Algorithm}&\textbf{Precision}&\textbf{Recall}&\textbf{Macro-F1} \\
    \midrule
		\multirow{1}{*}{1}& \textbf{DTW} & 15.47\% & 9.63\% & 53.23\% \\
		\hline
		\multirow{1}{*}{2} & \textbf{FC-LSTM} &13.29 \% & 52.27 \% & 53.78 \%\\
		\hline
		\multirow{3}{*}{3}& \textbf{DCRNN} & 13.97 \% & 57.81 \% & 54.42 \%\\
		& \textbf{STGCN} & 16.35 \% & 52.53 \% & 56.29 \%\\
		& \textbf{LRGCN} & 17.38 \% & 61.34 \% & 57.70 \% \\
		\hline
		\multirow{2}{*}{4} & \textbf{LRGCN-SAPE (static)} & 17.67 \% & \textbf{65.28} \% & 60.55 \%\\
		& \textbf{LRGCN-SAPE (evolving)} & \textbf{19.23} \% & 65.07 \% & \textbf{61.89} \%\\
	  \bottomrule
\end{tabular}
}
\eat{\vspace{-0.4cm}}
\end{table}

\subsubsection{Effect of the number of views}
\begin{figure}
\begin{center}
\includegraphics [width=0.37\textwidth,scale=1]{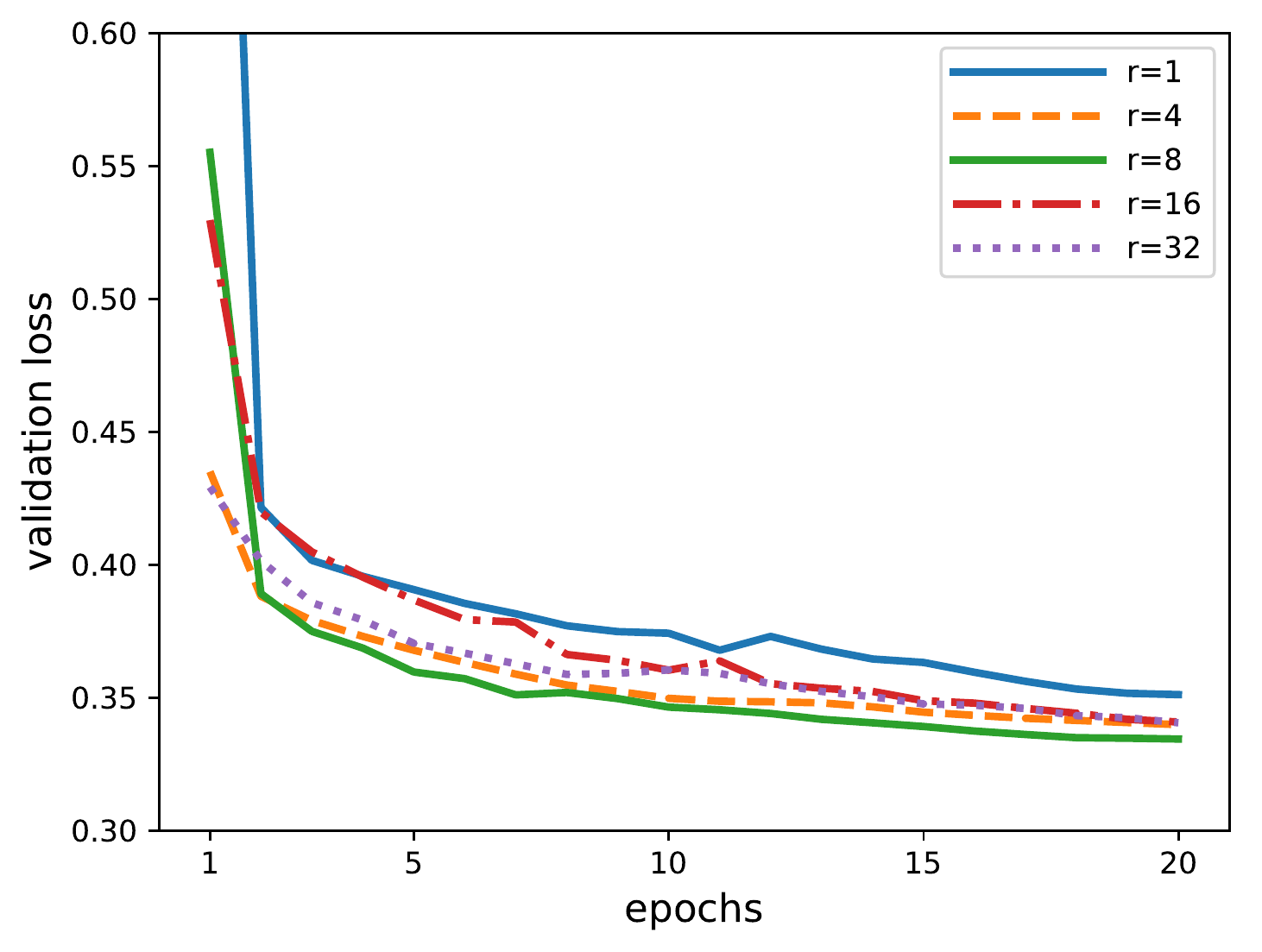}
\end{center}

\caption{Effect of the number of views (r) in SAPE on Traffic}
\label{fig.va}
\end{figure}

We evaluate how the number of views ($r$) affects the model performance.  Taking LRGCN-SAPE (static) on Traffic as an example, we vary $r$ from 1 to 32 and plot the corresponding validation loss with respect to the number of epochs in Figure \ref{fig.va}.  As we increase $r$, the performance improves (as the loss drops) and achieves the best when $r = 8$ (the green line).  We also observe that the performance difference for different $r$ is quite small, which demonstrates that our model performs very stably with respect to the setting of $r$.

\begin{table}
  \caption{Comparison of different methods on path congestion prediction on Traffic}
  \label{traffic}
  \scalebox{0.9}{
  \begin{tabular}{ccccc}
    \toprule
    &\textbf{Algorithm}&\textbf{Precision}&\textbf{Recall}&\textbf{Macro-F1} \\
    \midrule
		\multirow{1}{*}{1}& \textbf{DTW} & 12.05\% & 39.12\% & 51.62\% \\
		\hline
		\multirow{1}{*}{2} & \textbf{FC-LSTM} &54.44 \% & 87.97 \% & 76.55 \%\\
		\hline
		\multirow{3}{*}{3}& \textbf{DCRNN} & 63.05 \% & \textbf{88.55} \% & 82.60 \%\\
		& \textbf{STGCN} & 64.52 \% & 86.15 \% & 82.41 \%\\
		& \textbf{LRGCN} & 65.15 \% & 87.65 \% & 83.74 \% \\
		\hline
		\multirow{2}{*}{4} & \textbf{LRGCN-SAPE (static)} & 67.74 \% & 88.44\% & 84.84 \%\\
		& \textbf{LRGCN-SAPE (evolving)} & \textbf{71.04} \% & 88.50 \%& \textbf{86.74} \%\\
	  \bottomrule
\end{tabular}
}
\vspace{-0.4cm}
\end{table}

\subsubsection{Path embedding visualization}

\begin{figure}
\begin{center}
\includegraphics [width=0.35\textwidth,scale=1]{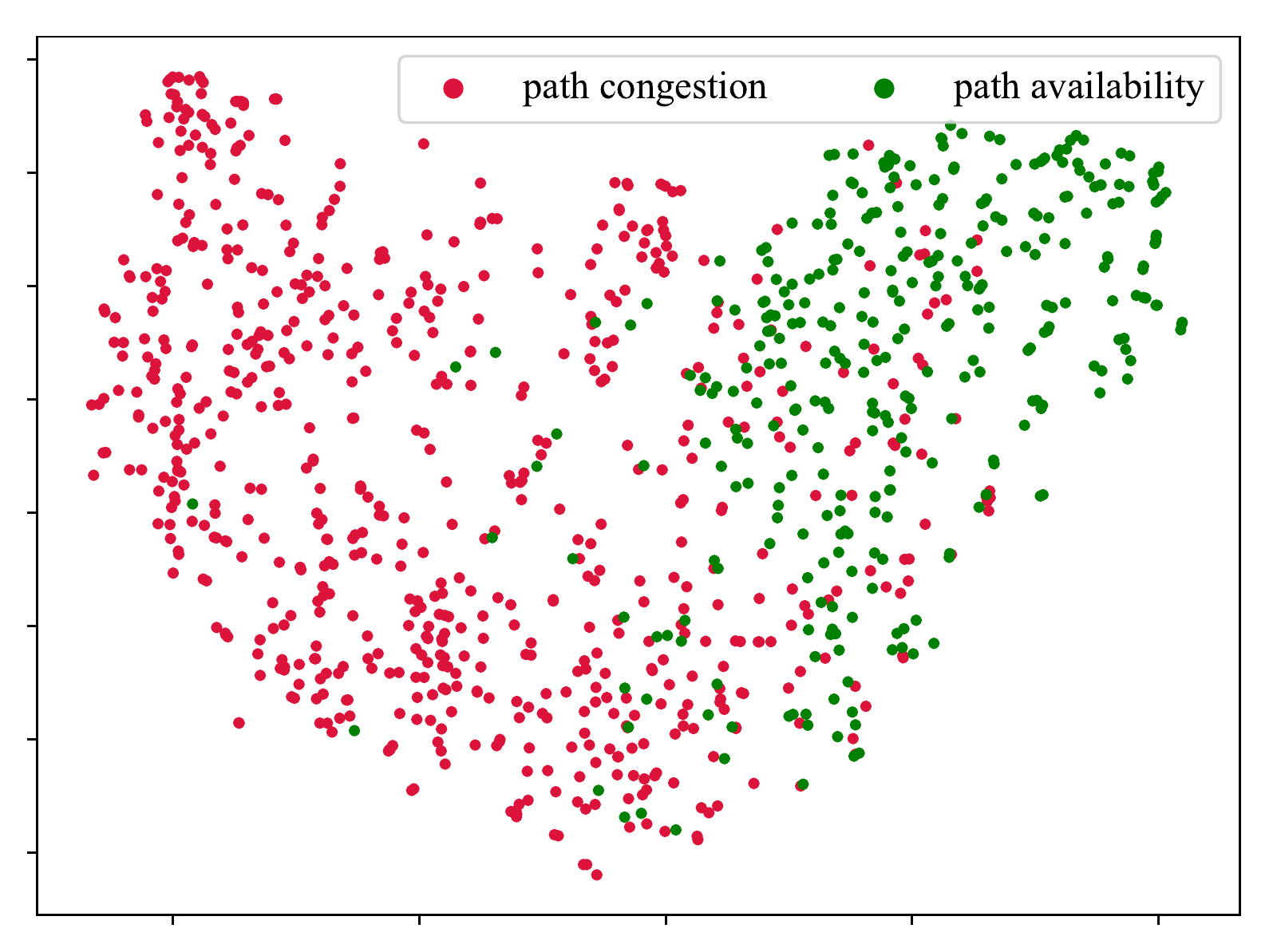}
\end{center}

\caption{Two-dimensional visualization of path embeddings on Traffic using SAPE.  The nodes are colored according to their path labels.}
\label{fig.diary}
\end{figure}
To have a better understanding of the derived path embeddings, we select 1000 path instances from the test set of Traffic data.  We apply LRGCN-SAPE (evolving) and derive the embeddings of these 1000 testing instances.  We then project the learned embeddings into a two-dimensional space by t-SNE \cite{maaten2008visualizing}, as depicted in Figure \ref{fig.diary}.  Green color represents path availability and red color represents path congestion.  As we can see from this two-dimensional space, \eat{the geometric distance between the path instances can reflect their path similarity properly.}paths of the same label have a similar representation, as reflected by the geometric distance between them.

\subsubsection{Effect of the normalization methods}
We compare the performance of asymmetric normalization $D^{-1}A$ and symmetric normalization $D^{-\frac{1}{2}}AD^{-\frac{1}{2}}$ on both Telecom and Traffic data sets in the LRGCN-SAPE (evolving) method.  Other experimental settings remain the same with the experiments presented above.  Results are listed in Table \ref{tab:nor}. For both data sets, asymmetric normalization outperforms symmetric normalization in both Precision and Macro-F1.  The advantage of asymmetric normalization is more significant on Telecom than on Traffic.  The reason is that most of the derived sensor stations in the traffic network are connected bidirectionally, while the switches in the telecommunication network are connected unidirectionally.  This demonstrates that asymmetric normalization is more effective than symmetric normalization on directed graphs.

\begin{table}
  \caption{Comparison of different normalization methods}
  \label{tab:nor}
  \scalebox{0.78}{
  \begin{tabular}{ccccccc}
    \toprule
Normalization&\multicolumn{3}{c}{\textbf{Telecom}}&\multicolumn{3}{c}{\textbf{Traffic}}\\
\hline
       -      &Precision&Recall&Macro-F1&Precision&Recall&Macro-F1\\
    \midrule
	$D^{-\frac{1}{2}}AD^{-\frac{1}{2}}$ &15.39\%&59.07\%&56.40\%&67.60\%&89.14\%&85.19\%\\
	$D^{-1}A$ &19.23\%&65.07\%&61.89\%&71.04\%&88.50\%&86.74\%\\
  \bottomrule
\end{tabular}
}
\vspace{-0.4cm}
\end{table}

\section{Related Work}\label{sec.related}
\eat{This work is related to prediction tasks in time-evolving graphs, path representation and neural network on graphs.}

Many real-world problems can be formulated as prediction tasks in time-evolving graphs.  We survey on two \eat{widely studied }tasks: failure prediction in telecommunication and traffic forecasting in transportation.  For failure prediction, the pioneer work \cite{klemettinen1999rule} formulates this task as a sequential pattern matching problem and the network topological structure is not fully exploited.  Later \cite{fronza2013failure} formulates it as a classification problem and uses SVM \cite{hearst1998support} to distinguish failures from normal behaviors.  \cite{pitakrat2018hora} makes use of Bayesian network to model the spatial dependency and uses Auto-Regressive Integrated Moving Average model (ARIMA) to predict node failures.  For traffic forecasting, existing approaches can be generally categorized into two groups.  The first group uses statistical models \cite{vlahogianni2014short}, where they either impose a stationary hypothesis on the time series or just incorporate cyclical patterns like seasonality.  Another group, on the other hand, takes advantage of deep learning neural networks to tackle non-linear spatial and temporal dependency.  \cite{Laptev2017TimeseriesEE} uses RNN to capture dynamic temporal dependency. \cite{zhang2017deep} applies CNN to model the spatial dependency between nodes.  All the above studies treat the underlying graph as a static graph, while our problem setting and solution target time-evolving graphs.  In addition, we study path failure prediction instead of node failure prediction.

Although there are many studies on node representation learning~\cite{perozzi2014deepwalk}, path representation has been studied less.  Deepcas ~\cite{li2017deepcas} leverages bi-directional GRU to sequentially take in node representation of a path forwardly and backwardly, and represents the path by concatenating the forward and backward hidden vectors.  ProxEmbed ~\cite{LiuZZZCWY17} uses LSTM to read node representation and applies max-pooling operation on all the time step outputs to generate the final path representation.  However when it comes to a long path, RNN may suffer from gradient exploding or vanishing, which prohibits the derived representation from reserving long-term dependency.  In this sense, our proposed path representation method SAPE utilizes the self-attentive mechanism, previously proven to be successful in \cite{DBLP:journals/corr/LinFSYXZB17,jiawww19}, to explicitly encode the node importance into a unified path representation.

There are many studies about\eat{in the area of} neural network on static graphs \cite{kipf2017semi,schlichtkrull2018modeling}. However, research that generalizes neural network to time-evolving graphs is still lacking.  The closest ones are neural networks on spatio-temporal data where graph structure does not change.  DCRNN \cite{li2018diffusion} models the static structure dependency as a diffusion process and replaces the matrix multiplications in GRU with the diffusion convolution to jointly handle temporal dynamics and spatial dependency.  STGCN \cite{yu2018spatio} models spatial and temporal dependency with three-layer convolutional structure, i.e., two gated sequential convolution layers and a\eat{ spatial} graph convolution layer in between.  Our solution LRGCN is novel as it extends neural networks to handle time-evolving graphs where graph structure changes over time.

\section{CONCLUSION}\label{sec.con}

In this paper, we study path classification in time-evolving graphs.  To capture temporal dependency and graph structure dynamics, we design a new neural network LRGCN, which views node correlation within a graph snapshot as intra-time relations, and views temporal dependency between adjacent graph snapshots as inter-time relations, and then jointly models these two relations.  To provide interpretation as well as enhance performance, we propose a new path representation method named SAPE.  Experimental results on a real-world telecommunication network and a traffic network in California show that LRGCN-SAPE outperforms other competitors by a significant margin in path failure prediction.  It also generates meaningful interpretations of the learned representation of paths.


\begin{acks}
The work described in this paper was supported by grants from the Research Grants Council of the Hong Kong Special Administrative Region, China [Project No.: CUHK 14205618], and Huawei Technologies Research and Development Fund.
\end{acks}

\bibliographystyle{ACM-Reference-Format}
\balance
\bibliography{sample-bibliography}

\clearpage
\appendix
\section{Data sets}\label{a.c}

\subsection{Telecommunication network}
The telecommunication data set is provided by a large telecommunication company and records sensor data from a real service session.
\begin{enumerate}
  \item \textbf{Data preparation} We choose a metropolitan LTE transmission network as our target which contains 626 switches and records sensor data from Feb 1, 2018 to Apr 30, 2018.
  \item \textbf{Static graph} Switches are linked by optical fiber in a directed way.  We treat a switch as a node in the graph and add a directed edge with weight $1$ between two nodes if there is an optical fiber linking one switch to another.  The statistics of the constructed static graph are listed in Table \ref{tab:sqq}.
  \begin{table}
  \caption{Statistics of constructed static graphs}
  \label{tab:sqq}
  \begin{tabular}{cccc}
    \toprule
    \textbf{Data set}&\textbf{Nodes}&\textbf{Edges}&\textbf{Density} \\
    \midrule
	Telecom&626&2464&0.63\%\\
	Traffic&4438&8996&0.05\%\\
  \bottomrule
\end{tabular}

\eat{\vspace{-0.3cm}}
\end{table}

\begin{table}
  \caption{A multivariate time series example recorded by switches}
  \label{tab:ss}
  \scalebox{0.9}{
  \begin{tabular}{cccccc}
    \toprule
    \textbf{SwitchID}&\textbf{Time}&\textbf{Sending power}&\textbf{Receiving power} \\
    \midrule
	H0001&20180201 00:00&-0.7 dB&20.7 dB\\
	H0001&20180201 00:15&-8.0 dB&18.1 dB\\
  \bottomrule
\end{tabular}

}
\eat{\vspace{-0.3cm}}
\end{table}
  \item \textbf{Feature matrix} Each switch records several observations every 15 minutes.  Among these observations, we use the average sending power and average receiving power as features.  For each switch, the sequence of features over time is a multivariate time series.  We exemplify a time series fragment in a 30-minute window in Table \ref{tab:ss}.  We normalize each feature to the range of [0, 1].  Finally we get a $8449 \times 626 \times 2$ feature matrix (8449 corresponds to the number of time steps).
  \item \textbf{Path labeling} There are 853 paths serving various services in this metropolitan transmission network.  An alarm system serves as an anomaly detector.  Once a path fault (e.g., network hardware outages, high transmission latency, or signal interference, etc.) is detected, it issues a warning message. If the number of warning messages within an hour exceeds a threshold, we label it as ``path failure''.  We use 24 hours' history data to predict if a path will fail in the next 24 hours.  Finally we get a $8449 \times 853$ label matrix.
  \item \textbf{Time-evolving graph} For an optical link, it can be categorized into two status: failure and availability, based on a key performance index called bit error rate.  We construct the time-evolving graph as follows: at time step $t$, we set the adjacency matrix element $A_{ij}^t=0$ if edge $\left<v_j, v_i\right>$ is labeled failure, and set $A_{ij}^t=1$ otherwise.
\end{enumerate}
\eat{
\begin{figure*}
\centering
\includegraphics [width=0.4\textwidth]{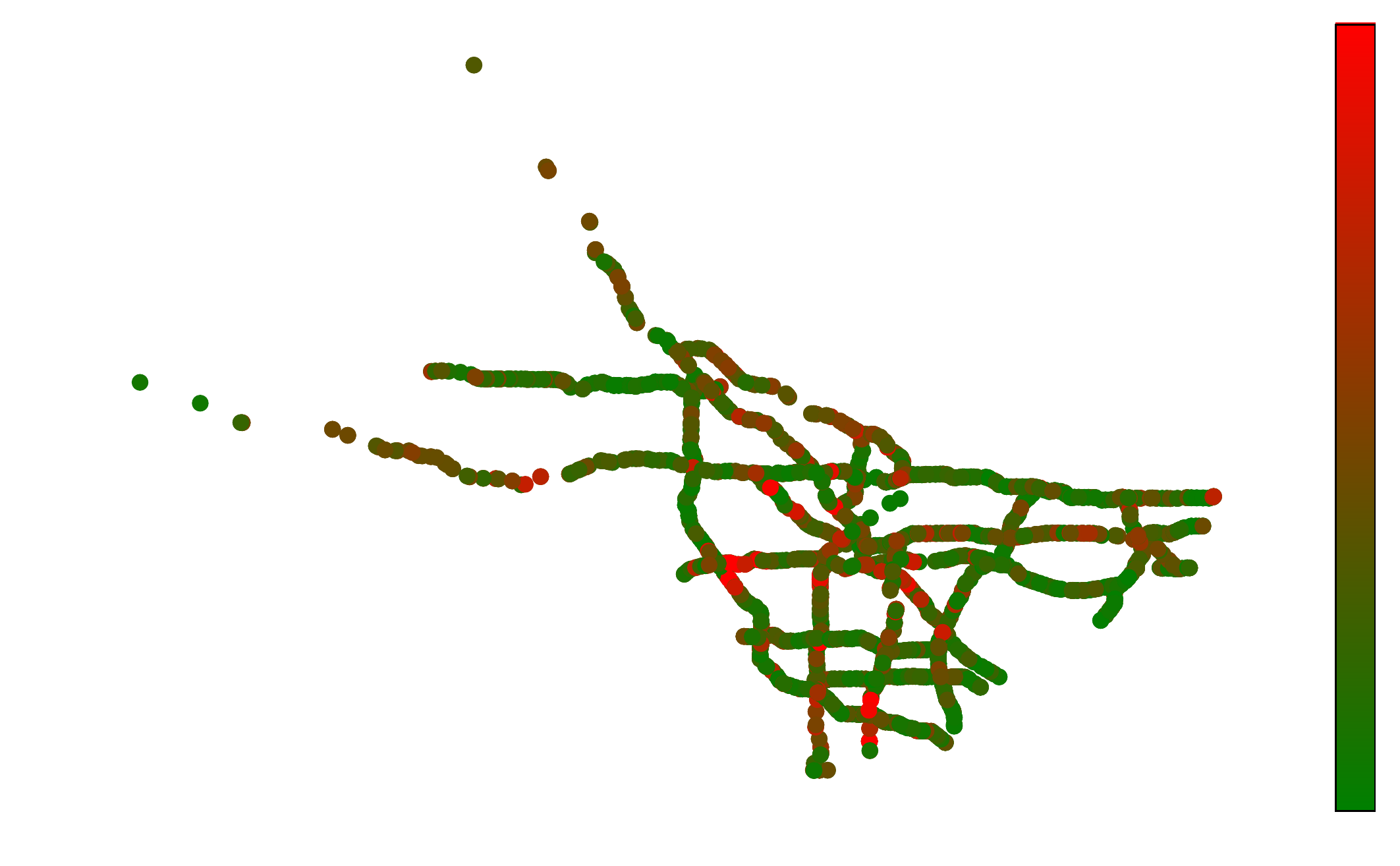}
\includegraphics [width=0.4\textwidth]{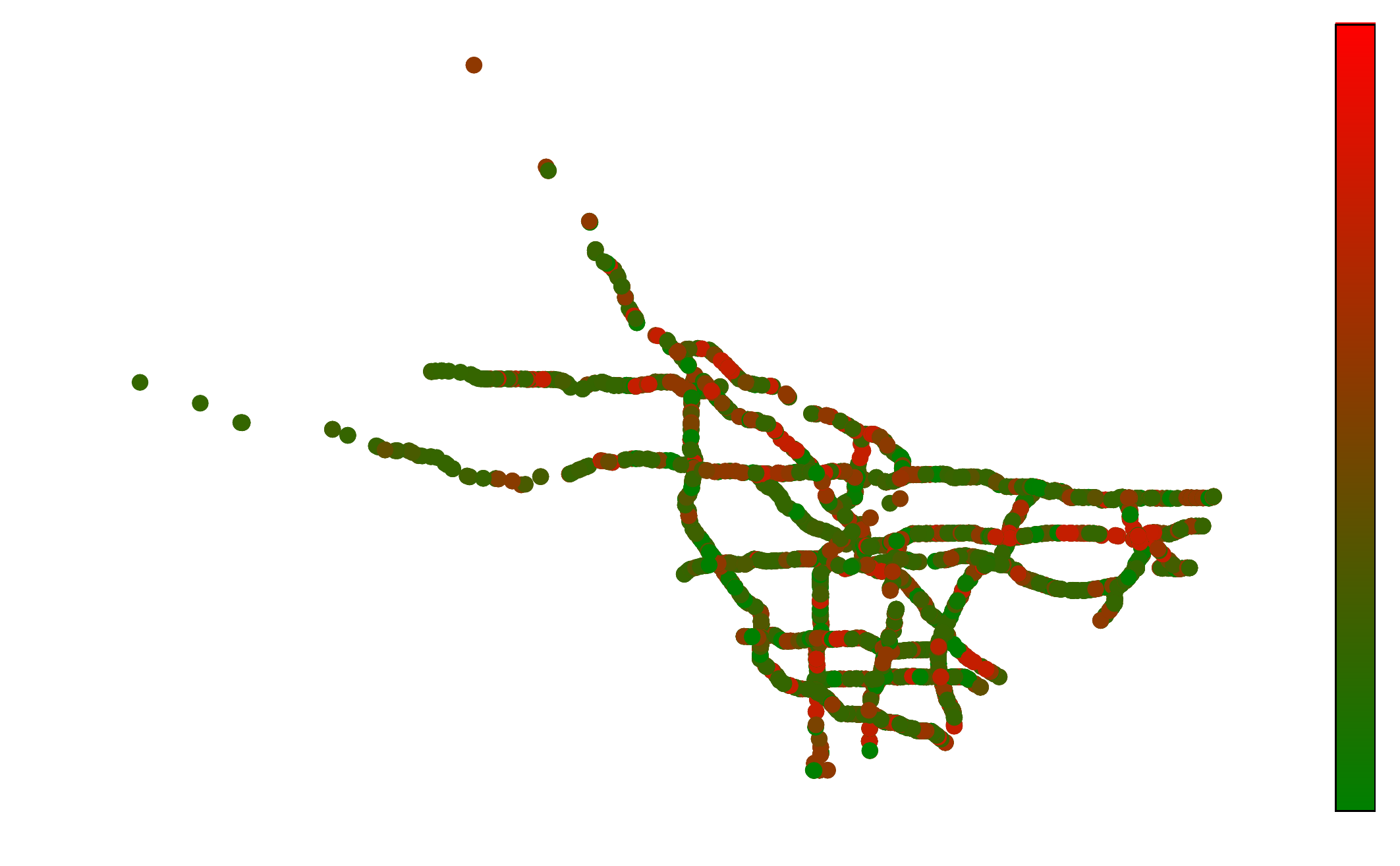}
\caption{Visualization of LRGCN (left) and DCRNN (right).  Red color implies probability of congestion predicted by the methods while green color represents predicted probability of light traffic.  By capturing structure dynamics such as edge removal, LRGCN acts as deblurring for some critical nodes.}
\label{comp}
\end{figure*}
}
\subsection{Traffic network}
\eat{We use the traffic data in District 7 for traffic congestion prediction. The details of data preprocessing is following.}
The traffic data set is collected by California Transportation Agencies (CalTrans) Performance Measurement System (PeMS).  The details can be found on the website of CalTrans\footnote{\url{http://pems.dot.ca.gov/}}.
\begin{enumerate}
  \item \textbf{Data preparation} We use the traffic data in District 7 from Jun 1, 2018 to Aug 30, 2018.  There are two kinds of information: \emph{real-time traffic} and \emph{meta data}.  The former records the traffic information at sensor stations such as the average speed, average occupancy, etc.  The latter records the general information of the stations such as longitude, latitude, delay, closure, etc.  The number of stations used in this study is 4438.
  \item \textbf{Static graph} In the \emph{meta data}, the order of the stations along the freeway is indicated by the ``Absolute Postmile'' field.  We treat a station as a node in the graph and connect the adjacent stations on a freeway along the same direction one by one (in ``Absolute Postmile'' order). The weight of every edge is set to 1.
  \item \textbf{Feature matrix} Each station records several kinds of traffic information hourly.  Among them, we use the average speed and average occupancy as features.  We replace missing values with 0, and normalize the features to the range of [0, 1].  Finally we get a $2160 \times 4438 \times 2$ feature matrix (2160 corresponds to the number of time steps).
  \item \textbf{Path labeling} We randomly choose two nodes on the static graph as the source and target, then use Dijkstra's algorithm to find the shortest path from the source to target. We sample 200 paths and restrict the path length to the range of [2, 50]. For a node, its congestion information is indicated by the ``delay'' field. For a path, if two consecutive nodes have congestion at the same time, we label it as ``path congestion''.  We use 24 hours' history data to predict if a path will congest in the next one hour\eat{, if yes we label it 1, otherwise 0}. Finally we get a $2160 \times 200$ label matrix.
  \item \textbf{Time-evolving graph} We construct the time-evolving graph according to the following rules.
\begin{itemize}
\item  At time step $t$, if $v_i$ is labeled closure, we delete all its incoming/outgoing edges in the graph snapshot at $t$.
\item  At time step $t$, if $v_i$ is labeled congestion, we shrink all its incoming/outgoing edge weights by a factor of 0.5.
\end{itemize}
\end{enumerate}

\section{Detailed Experiment Settings}\label{a.b}
This part details the implementation of each method and their hyperparameter setting if any.

\noindent\textbf{DTW}: Dynamic Time Warping is an algorithm for measuring similarity between two time series sequences.  As the time series on each node is multivariate, we calculate the sum of squared DTW similarities for all variables.  We consider a path as a bag of nodes, and calculate the similarity between paths by their maximum node similarity. In the DTW method, we do not model node correlations, which means graph structure is not taken into consideration.


\noindent\textbf{FC-LSTM} uses two-layer LSTM neural networks for modeling temporal dependency, another LSTM layer for path representation and a fully connected layer.  In the two-layer LSTM, the first layer is initialized with zero, and its last hidden state is used to initialize the second LSTM layer. The output dimension of these two LSTM is 8.  The LSTM path representation layer is used to derive a fixed-length path representation. It works as follows:
\begin{itemize}
\item Indexing node representations of a path from the last hidden state of the previous LSTM.
\item Feeding this hidden representation sequence to a LSTM layer.
\item The last hidden state of this LSTM is the final path representation.
\end{itemize}
The output dimension of this LSTM is also 8. FC-LSTM does not model node correlations, and it can be regarded as LRGCN with $A = I_N$.

\noindent\textbf{DCRNN} uses two-layer DCRNN for static graph modeling, another LSTM layer for path representation and a fully connected layer.  The difference between DCRNN and FC-LSTM is that the former models the node correlation as a diffusion process while the latter does not consider node correlation. For parameters, its maximum diffusion step is 3 and the output dimensions of both DCRNN and LSTM are 8.

\noindent\textbf{STGCN} uses two-layer STGCN for static graph modeling, another LSTM layer for path representation and a fully connected layer.  STGCN models node correlation and temporal dependency with three-layer convolutional structure, i.e., two convolution layers and one GCN layer in between.  For parameters, the graph convolution kernel size is set to 1 and the temporal convolution kernel size is set to 3.  The output dimensions of both STGCN and LSTM are 8.

\noindent\textbf{LRGCN} is the same as DCRNN except that we replace the first two-layer DCRNN with LRGCN. The hidden dimension $h$ is 96.

\noindent\textbf{LRGCN-SAPE (static)}: The difference between this method and LRGCN is that the path representation is derived by SAPE instead of LSTM.  For parameters of SAPE, we set $v = 8$, $d_s = 32$ and $r = 8$.

\noindent\textbf{LRGCN-SAPE (evolving)}: The main advantage of LRGCN is that it can model time-evolving graphs.  In this method, graph structure dynamics are modeled. The parameters of SAPE are set the same as the above method.

\eat{

\section{Comparison between LRGCN AND DCRNN}
In this supplementary experiment, we make a comparison between LRGCN and DCRNN through a node classification task on Traffic data.  In method LRGCN (evolving), we construct a one-layer LRGCN on a time-evolving graph.  The output dimension of LRGCN is 1. By applying a sigmoid function, the output of LRGCN in the last time step can be regarded as the congestion probability of each node.  For the comparison method DCRNN (static), we construct a one-layer DCRNN on a static graph.  The output dimension of DCRNN is 1.  We apply a sigmoid function and get the congestion probability of each node.  We visualize the results for the same prediction time interval in Figure \ref{comp}.  We find that the visual effect of DCRNN (static) is a smooth blurring, which makes the color stretch over its neighborhood and become less distinct, while LRGCN (evolving) can retain the distinctive color of some critical nodes.  The reason is as follows: DCRNN (static) always smoothes node representations over the graph structure regardless of the graph structure dynamics.  In contrast, LRGCN (evolving) can $\emph{\textbf{deblur}}$ some critical nodes (e.g., nodes near accidents) by capturing the graph structure dynamics, e.g., node closures, thus this deblurred node representation can be used for downstream classification tasks.
}

\end{document}